\documentclass[11pt]{article}
\usepackage[preprint]{acl}
\usepackage[english]{babel}
\usepackage{microtype}
\usepackage{amsmath,amssymb}
\usepackage{graphicx}
\usepackage{booktabs}
\usepackage{tabularx}
\usepackage{multirow}
\usepackage{enumitem}
\usepackage{xcolor}
\usepackage{colortbl}
\usepackage{arydshln}
\usepackage{float}
\usepackage{xspace}
\usepackage{balance}

\definecolor{gainblue}{HTML}{D9EAF7}
\definecolor{tiegray}{HTML}{F2F2F2}
\definecolor{lossorange}{HTML}{FCE4D6}
\definecolor{lossred}{HTML}{F4CCCC}

\newcommand{\chrf}{ChrF2\xspace}
\newcommand{\kiwixxlfull}{CometKiwi-XXL}
\newcommand{\kiwixxl}{Kiwi\textsubscript{XXL}}
\newcommand{\metricxfull}{MetricX-24-XXL}
\newcommand{\metricx}{Mx24\textsubscript{XXL}}
\newcommand{\metricxref}{Mx24\textsubscript{XXL}\textsuperscript{ref}}
\newcommand{\metricxqe}{Mx24\textsubscript{XXL}\textsuperscript{QE}}
\newcommand{\xcometxl}{XCOMET-XL\xspace}
\newcommand{\pGain}[1]{\cellcolor{gainblue}#1}
\newcommand{\pTie}[1]{\cellcolor{tiegray}#1}
\newcommand{\pLoss}[1]{\cellcolor{lossorange}#1}
\newcommand{\pStrong}[1]{\cellcolor{lossred}\textbf{#1}}
\usepackage[htt]{hyphenat}   
\title{Quantize, Qualify, Rerank: A Recipe for Compressing LLMs Without Losing Quality}

\author{Thamme Gowda \\
  Microsoft \\
  \texttt{thammegowda@microsoft.com}}

\begin{document}
\maketitle

\begin{abstract}
Compressing large language models lowers serving cost, but a single automatic
metric may miss quality loss. For the WMT26 Model Compression constrained track,
we compress \texttt{google/gemma-3-12b-it} ($24.4$\,GB in bf16) for three
translation directions on one H100. Paired bootstrap analysis exposes metric
disagreement on 4-bit quantization; guided by WMT24 human meta-evaluation, we use
\metricxfull{} as primary and \kiwixxlfull{} as a cross-check. Removing the unused vision
stack and non-task vocabulary and using int8 weights or fp8 activations causes no
resolved loss, whereas int4 leaves a small residual. QLoRA healing overfits its
calibration domain. A Cometoid best-of-4 reranker at $T{=}0.3$ improves both the
8-bit and 4-bit systems under WMT25 test set. On post-submission WMT26,
both remain competitive with bf16 under reference-free metrics; the small
difference between the reranked variants is treated as noise.
\end{abstract}

\section{Introduction}
\label{sec:intro}

General-purpose large language models (LLMs) spend most of their capacity on broad competence. A machine translation (MT)
deployment restricted to three fixed language pairs asks a different question:
which parts can be removed, quantized, or replaced without a user noticing? We treat
``without noticing'' as a measurement problem rather than infer it from a raw metric
ordering. We first compare several metrics with paired bootstrap resampling; when
they disagree, we use WMT Metrics Shared Task's meta-evaluation as an external anchor. Methods
that learn from examples, including calibration, adapter tuning, pruning recovery,
and QLoRA-style adaptation \citep{dettmers2023qlora}, must retain gains across
domains.

No MT metric is uniformly reliable across model and output regimes, particularly
for close systems. WMT therefore meta-evaluates evolving metric panels against
human judgments each year \citep{callison-burch-etal-2008-meta,
freitag-etal-2024-llms,lavie-etal-2025-findings}. When our local metrics disagree,
the WMT24 human ranking supplies study-specific external evidence. We test
data-dependent compression under domain shift.

We study this question in the WMT26 Model Compression constrained
track,\footnote{\url{https://www2.statmt.org/wmt26/model-compression.html}}
specializing the prescribed Gemma-3-12B model for three translation directions.
We evaluate compression levers alone and in composition, then test whether their
apparent gains survive metric qualification, held-out domains, and the post-submission
WMT26 blind set.

\paragraph{Contributions and findings.}
\begin{enumerate}[nosep,leftmargin=1.4em]
\item \textbf{Metric qualification.} We show that many popular metrics can miss
or disagree on quantization losses, and present a paired-bootstrap qualification
procedure that measures score-ordering stability and anchors metric choice in
human meta-evaluation (Section~\ref{sec:protocol}).
\item \textbf{Data-free compression.} We combine vision/vocabulary removal,
int8 embeddings, and int8/fp8 quantization without calibration, and show
that qualified metrics resolve no 8-bit loss, while int4 retains a small residual
(Sections~\ref{sec:vocab},~\ref{sec:quant}).
\item \textbf{Generalization gate.} We show that QLoRA healing can recover in its
calibration domain while substantially degrading quality out of domain, motivating
held-out-domain validation for learned compression (Section~\ref{sec:prune}).
\item \textbf{Inference-time recovery.} Best-of-$k$ reranking improves both
compressed systems under held-out WMT25 metrics and remains competitive with
bf16 on WMT26; small differences between reranked variants are treated as noise
(Section~\ref{sec:inference}).
\end{enumerate}

\section{WMT26 Model Compression Task}
\label{sec:setup}

The constrained track compresses \texttt{google/gemma-3-12b-it} \citep{gemma3}
for Czech$\to$German, English$\to$Simplified Chinese, and English$\to$Egyptian
Arabic on one H100. Systems are ranked on a quality$\times$size$\times$speed
Pareto frontier; size includes disk and peak load-time VRAM, and speed is
end-to-end tokens/s.\footnote{\url{https://github.com/thammegowda/wmt26-model-compression}}
We used the 1121-segment WMT25 set for development and scored WMT26 only after
submissions were fixed.

We serve the text path with \emph{tahoma}, a C++ runtime with native int4/int8
\citep{machete} and fp8 kernels. Its serving architecture follows vLLM
\citep{kwon2023vllm}, including paged attention, continuous batching, and CUDA
graph replay.

\section{Compression Levers}
\label{sec:worked}

Compression choices should follow parameter mass: MLPs account for 72\% of the
text model, attention for 19\%, and the tied embedding/LM-head table for 9\%.
We consider \emph{structural} removal and \emph{representational} precision
changes, evaluating data-free levers before calibration-dependent methods.

\subsection{Structural: Removing Deadweight}
\label{sec:vocab}

\paragraph{The vision stack.} The base checkpoint is multimodal, but three
translation directions never see an image. Its vision stack, a $417$M-parameter
SigLIP encoder plus multimodal projector occupying $0.84$\,GB in bf16, is therefore unused.
We drop it outright, data-free, before any other lever; the $23.5$\,GB baseline used
throughout is this text-only model. The tokenizer contains further removable
capacity.

\paragraph{Vocabulary.} The Gemma tokenizer carries 262{,}208 pieces covering every language and code;
three translation directions need a handful of scripts. We categorize every piece
by the Unicode \emph{script} of its letters and keep only system/special tokens,
the 256 byte-fallback tokens, and pieces whose letters lie in the task scripts
(Latin, Arabic, Han); digits, punctuation, and symbols are script-neutral and
allowed. As an additional rule, long symbol composites ($\geq3$ symbol characters,
e.g.\ code/markup fragments) are dropped while short ones are kept.

This drops 66{,}831 pieces (25.5\%) from non-task scripts. It is
safe because any dropped piece still round-trips through byte fallback. The
embedding table shrinks from 2.01\,GB to 1.50\,GB (bf16), and composes with int8
embeddings for a further reduction (Table~\ref{tab:vocab}). 

\begin{table}[t]
\centering\small
\setlength{\tabcolsep}{3.5pt}
\begin{tabular}{lrr}
\toprule
\textbf{Stage} & \textbf{Pieces} & \textbf{Embedding (bf16)} \\
\midrule
Full tokenizer & 262{,}208 & 2.01\,GB \\
Script keep-set & 195{,}377 & 1.50\,GB \\
\;\;+ int8 embeddings & 195{,}377 & ${\sim}0.75$\,GB \\
\bottomrule
\end{tabular}
\caption{Vocabulary specialization: 25.5\% fewer pieces with byte-fallback
coverage.}
\label{tab:vocab}
\end{table}

\subsection{Representational: Precision~Variants}
\label{sec:quant-variants}

Beyond removing deadweight, we compare the baseline bf16 checkpoint against six
alternative numeric representations of the same weights
(Table~\ref{tab:precision-variants}). All variants except NF4 run on \emph{tahoma};
NF4 is bitsandbytes' learned-codebook 4-bit quantization, applied online through
vLLM rather than tahoma, making it a cross-engine reference.

\begin{table}[t]
\centering\small
\begin{tabular}{lll}
\toprule
\textbf{System} & \textbf{W/A} & \textbf{Role} \\
\midrule
bf16  & bf16/bf16 & native baseline \\
fp32  & fp32/fp32 & near-null probe \\
i8a16 & int8/bf16 & weight-only, $2\times$ \\
f8a8  & fp8/fp8   & speed lever \\
i4a16 & int4/bf16 & weight-only, $3.5\times$ \\
i4a8  & int4/fp8  & smallest+fastest 4-bit \\
NF4   & FP4/bf16 & codebook, cross-engine \\
\bottomrule
\end{tabular}
\caption{Precision variants compared throughout the paper (weights/activations
format). Section~\ref{sec:protocol} qualifies which metrics resolve differences
among them; Section~\ref{sec:quant} reports their measured scores.}
\label{tab:precision-variants}
\end{table}

The baseline checkpoint is natively bf16, so we use it as the reference; fp32
accumulation is an engineering near-null probe, and fp16 is excluded because it
overflows Gemma-3 activations. The i4a16 recipe uses asymmetric blockwise
quantization with MSE-optimal clipping. We combine the same int4 weights with fp8
activations in a W4A8 kernel (i4a8), whose matmuls use Hopper fp8 tensor cores.

Blockwise 4-bit scales trade storage for granularity: a smaller block gives finer
scales but stores more of them. The deployable window is set by kernel geometry,
not preference: each block must divide every quantized weight's input dimension
$K\!\in\!\{3840,4096,15360\}$, whose GCD is $256$, so $256$ is the largest
admissible block and $384/512/768$ are runtime-rejected (e.g.\ $384\nmid4096$);
below $64$, block $32$ is refused by the Machete W4A16 kernel, which supports only
group sizes ${\ge}64$. This leaves $\{64,128,256\}$ as the only deployable block
sizes; Section~\ref{sec:quant} reports their measured quality
(Table~\ref{tab:blocksize}).

\section{Metric Qualification and Bootstrap Analysis}
\label{sec:metrics}
\label{sec:protocol}

We compare \chrf{}~\citep{popovic2015chrf,post2018sacrebleu} under both
mean-segment and corpus aggregation. We compute
COMET22~\citep{rei2020comet}, BLEURT20~\citep{sellam2020bleurt},
CometKiwi22~\citep{rei2022cometkiwi}, and \kiwixxlfull{}
(\kiwixxl{})~\citep{rei2023scaling}, and Cometoid~\citep{cometoid} with
PyMarian~\citep{pymarian}. We additionally score
\xcometxl{}~\citep{guerreiro-etal-2024-xcomet} and evaluate \metricxfull{}
(\metricx{})\footnote{MetricX implementation:
\url{https://github.com/thammegowda/metricx}.}~\citep{juraska2024metricx24} in
reference-based (\metricxref{}) and QE (\metricxqe{}) modes on the variants of
Section~\ref{sec:quant-variants}.

We run paired bootstrap resampling separately for each metric and system pair
\citep{koehn-2004-statistical}. For rows with additive segment scores, let $x_i$
be the signed score difference on segment $i$: bf16 minus the variant for
higher-is-better metrics, and the variant minus bf16 for lower-is-better
\metricx{}. Thus $x_i>0$ always favors bf16. Each replicate draws $N{=}1121$
values with replacement from the observed paired differences $x_1,\ldots,x_N$.
If $x_j^{(b)}$ is the $j$th value drawn in replicate $b$, we
compute
\begin{equation}
\begin{aligned}
 \bar x^{(b)}=&\frac{1}{N}\sum_{j=1}^{N}x_j^{(b)} \\
 S_M=&\frac{1}{B}\sum_{b=1}^{B}
  \mathbf{1}\!\left[\bar x^{(b)}>0\right]
\end{aligned}
 \label{eq:bootstrap-support}
\end{equation}
with $B{=}4000$ replicates. Here $\bar x^{(b)}$ is the mean difference in the $b$th
virtual test set, and $\mathbf{1}[\cdot]$ is one when its condition holds. Therefore
$S_M{=}97\%$ means that bf16 obtains the better mean metric score in 97\% of the
virtual test sets; it is not the probability of a human-quality loss. Values near
50\% leave the metric ordering unresolved; it does not sufficiently prove quality equivalence.

The \textit{ChrF2 (mean seg.)} row applies this protocol to sentence-level ChrF2
scores. For \textit{ChrF2 (corpus)}, each replicate instead uses the same paired
indices to compute SacreBLEU's corpus-level ChrF2 difference.

We estimate the required sample size as
\begin{equation}
 N^\star=\left(\frac{1.96\,s}{\bar x}\right)^2
 \label{eq:nstar}
\end{equation}
where $\bar x$ and $s$ are the observed mean and standard deviation of the signed
differences for that comparison. It estimates how many segments would be needed for a
two-sided 95\% interval to exclude zero if the observed effect and variance persist;
we show $\infty$ when $\bar x\leq0$ because the observed direction does not favor
bf16. This additive-segment extrapolation is undefined for the nonlinear
corpus-ChrF2 aggregation. We treat fp32/int8/fp8 as near-null probes and int4 as
the stronger perturbation.

\begin{table*}[t]
\centering
\renewcommand{\arraystretch}{1.1}
\setlength{\dashlinedash}{0.5pt}
\setlength{\dashlinegap}{1.4pt}
\textbf{Bootstrap support $S_M$ and one-sided $N^\star$}\\[-0.2em]
{\small
\setlength{\tabcolsep}{1.5pt}
\begin{tabular}{lrrr:rrrr:rrrrrr}
\toprule
\multicolumn{14}{c}{\colorbox{gainblue}{\strut $S_M\leq5\%$: variant wins in $\geq95\%$ of resamples}\quad
\colorbox{tiegray}{\strut $5\%<S_M<95\%$: neither wins in $\geq95\%$}} \\
\multicolumn{14}{c}{\colorbox{lossorange}{\strut $95\%\leq S_M<99\%$: bf16 wins in $95$--$99\%$ of resamples}\quad
\colorbox{lossred}{\strut $S_M\geq99\%$: bf16 wins in $\geq99\%$}} \\
\midrule
& & \multicolumn{2}{c}{\textbf{fp32}} & \multicolumn{2}{c}{\textbf{i8a16}} & \multicolumn{2}{c}{\textbf{f8a8}} & \multicolumn{2}{c}{\textbf{i4a16}} & \multicolumn{2}{c}{\textbf{i4a8}} & \multicolumn{2}{c}{\textbf{NF4}} \\
\cmidrule(lr){3-4}\cmidrule(lr){5-6}\cmidrule(lr){7-8}\cmidrule(lr){9-10}\cmidrule(lr){11-12}\cmidrule(lr){13-14}
\textbf{Metric} & \textbf{Size} & \textbf{$S_M$} & \textbf{$N^\star$} & \textbf{$S_M$} & \textbf{$N^\star$} & \textbf{$S_M$} & \textbf{$N^\star$} & \textbf{$S_M$} & \textbf{$N^\star$} & \textbf{$S_M$} & \textbf{$N^\star$} & \textbf{$S_M$} & \textbf{$N^\star$} \\
\midrule
CometKiwi22 & 565M & \pTie{$45.3\%$} & $\infty$ & \pTie{$22.7\%$} & $\infty$ & \pTie{$23.1\%$} & $\infty$ & \pTie{$73.0\%$} & 12.1k & \pTie{$10.7\%$} & $\infty$ & \pTie{$81.7\%$} & 5.5k \\
Cometoid & 569M & \pTie{$41.0\%$} & $\infty$ & \pTie{$74.7\%$} & 12.6k & \pTie{$30.0\%$} & $\infty$ & \pTie{$41.4\%$} & $\infty$ & \pTie{$6.6\%$} & $\infty$ & \pTie{$44.8\%$} & $\infty$ \\
BLEURT20 & 576M & \pGain{$2.8\%$} & $\infty$ & \pTie{$12.6\%$} & $\infty$ & \pTie{$49.0\%$} & $\infty$ & \pTie{$60.2\%$} & 68.1k & \pTie{$39.8\%$} & $\infty$ & \pGain{$4.9\%$} & $\infty$ \\
COMET22 & 581M & \pTie{$93.3\%$} & 2.0k & \pTie{$60.9\%$} & 87.1k & \pTie{$58.6\%$} & 103k & \pTie{$58.7\%$} & 85.1k & \pTie{$89.0\%$} & 2.8k & \pTie{$48.1\%$} & $\infty$ \\
\xcometxl{} & 3.49B & \pTie{$6.8\%$} & $\infty$ & \pTie{$13.2\%$} & $\infty$ & \pGain{$4.2\%$} & $\infty$ & \pTie{$35.3\%$} & $\infty$ & \pTie{$53.2\%$} & 570k & \pTie{$10.1\%$} & $\infty$ \\
\kiwixxl{} & 10.7B & \pTie{$65.4\%$} & 24.2k & \pTie{$55.4\%$} & 445k & \pTie{$21.0\%$} & $\infty$ & \pLoss{$98.2\%$} & 989 & \pTie{$56.0\%$} & 334k & \pStrong{$99.4\%$} & 711 \\
\metricxref & 12.9B & \pTie{$70.1\%$} & 14.1k & \pTie{$70.8\%$} & 16.0k & \pTie{$67.8\%$} & 23.0k & \pStrong{$100.0\%$} & \textbf{172} & \pStrong{$100.0\%$} & 337 & \pStrong{$100.0\%$} & 450 \\
\metricxqe & 12.9B & \pTie{$79.0\%$} & 5.8k & \pTie{$36.7\%$} & $\infty$ & \pTie{$65.7\%$} & 28.9k & \pStrong{$100.0\%$} & \textbf{184} & \pStrong{$100.0\%$} & \textbf{325} & \pStrong{$99.9\%$} & 500 \\
\shortstack {ChrF2(mean seg.)} & -- & \pTie{$62.2\%$} & 44.0k & \pTie{$67.4\%$} & 22.6k & \pLoss{$98.8\%$} & 892 & \pStrong{$100.0\%$} & 258 & \pStrong{$100.0\%$} & 196 & \pStrong{$100.0\%$} & 129 \\
\shortstack {ChrF2(corpus)} & -- & \pTie{$57.1\%$} &  & \pTie{$81.8\%$} &  & \pLoss{$98.2\%$} &  & \pStrong{$100.0\%$} &  & \pStrong{$100.0\%$} &  & \pStrong{$100.0\%$} &  \\
\bottomrule
\end{tabular}
}
\caption{Paired-bootstrap support $S_M$ for bf16 over each variant. $N^\star$
estimates the sample size needed to resolve a bf16-favoring effect at 95\%;
$\infty$ denotes the opposite observed direction. Corpus ChrF2 has no additive
$N^\star$.}
\label{tab:metric-protocol}
\end{table*}

The local study is ambiguous. Both \metricx{} modes strongly favor bf16 over the two
Tahoma int4 variants. \kiwixxl{} favors bf16 over i4a16, while i4a8 remains
unresolved; several other learned metrics do not resolve a consistent 4-bit effect.
\chrf{} also favors bf16.

\paragraph{External anchor.}
WMT24 ranks \metricx{} in the top tier against professional MQM/ESA judgments and
\kiwixxl{} above older baselines \citep{freitag-etal-2024-llms}; \chrf{} ranks much
lower. We therefore use \metricx{} as primary and \kiwixxl{} as a reference-free
cross-check, describing differences as metric residuals rather than demonstrated
human-significant losses.

\section{Effective Compression Levers}
\label{sec:quant}

\paragraph{Precision, engines, and blind-set transfer.}
Table~\ref{tab:trusted-scores} combines WMT25 development results with the WMT26
test set, kept blind during development. On both sets, i8a16 and f8a8 remain near
bf16, whereas i4a16, i4a8, and vLLM NF4 leave larger metric residuals; these are
not claims of human-perceptible loss. Both best-of-4 systems remain competitive
with bf16 on WMT26. Their differences are small and not significance-tested, so
we do not interpret i4a8-bok4's numerical lead as evidence that int4 is better.

The engine controls separate compression from implementation effects. tahoma's
bf16 quality is close to hf-bf16 across all four score columns\footnote{tahoma's
Gemma-3 originally closely reimplemented HuggingFace \texttt{transformers}, but
inadvertently omitted linear RoPE scaling, slightly degrading quality. The
omission was identified by cross-checking vLLM's reference implementation and
fixed; the post-fix scores are reported here, with hf and vLLM controls
unchanged.}, with small numerical differences across engines. vLLM bf16/fp8 changes
are mixed, so we do not interpret their apparent metric gains. vLLM quantizes
the bf16 snapshot on load; its FP8/NF4 disk values come from equivalent static
exports. At steady state, tahoma f8a8 matches vLLM FP8 ($6094$ vs.\
$5934$\,tok/s), while i4a8 exceeds vLLM NF4 ($5131$ vs.\ $3190$\,tok/s).
With the same local checkpoint, tahoma cold start is $24$--$34$\,s, versus
$55$--$168$\,s for vLLM FP8 and $92$\,s for NF4.

\paragraph{4-bit block size.}
Of the deployable sizes in Section~\ref{sec:quant-variants}, block 128 gives the
best quality (Table~\ref{tab:blocksize}); block 256 is slightly smaller and
faster but worse, so submissions use 128.

\begin{table}[t]
\centering\small
\setlength{\tabcolsep}{2pt}
\begin{tabular}{lrrrr}
\toprule
\textbf{Block} & \textbf{\kiwixxl{}}$\uparrow$ & \textbf{\metricxref}$\downarrow$ & \textbf{Size} & \textbf{tok/s} \\
\midrule
$64$  & $64.61$ & $5.89$ & $7.31$ & $4706$ \\
$128$ & $\mathbf{65.22}$ & $\mathbf{5.81}$ & $6.96$ & $4723$ \\
$256$ & $64.28$ & $6.11$ & $\mathbf{6.78}$ & $\mathbf{4758}$ \\
\bottomrule
\end{tabular}
\caption{Deployable 4-bit block sizes; size is on-disk GB.}
\label{tab:blocksize}
\end{table}

\paragraph{8-bit embeddings are nearly lossless.}
Per-row int8 quantization cuts the tied embedding/LM-head table from $2.01$ to
${\sim}1.0$\,GB with changes of at most $0.18$ \kiwixxl{} and $0.04$
\metricxref{} (Table~\ref{tab:embed}); all submissions use it.

\begin{table}[t]
\centering\small
\setlength{\tabcolsep}{3.5pt}
\begin{tabular}{lrr:rr}
\toprule
& \multicolumn{2}{c}{\textbf{\kiwixxl{}}$\uparrow$}
& \multicolumn{2}{c}{\textbf{\metricxref}$\downarrow$} \\
\cmidrule(lr){2-3}\cmidrule(lr){4-5}
\textbf{Variant} & \textbf{bf16} & \textbf{int8}
& \textbf{bf16} & \textbf{int8} \\
\midrule
i8a16 & 65.10 & 65.16 & 5.66 & 5.68 \\
i4a16 & 64.58 & 64.62 & 5.87 & 5.89 \\
i4a8  & 64.96 & 64.93 & 5.87 & 5.85 \\
f8a8  & 65.24 & 65.42 & 5.63 & 5.67 \\
\bottomrule
\end{tabular}
\caption{Per-row int8 embedding ablation.}
\label{tab:embed}
\end{table}

\subsection{Recovery via Calibration: Domain Sensitivity}
\label{sec:prune}

We QLoRA-heal i4a8 on GlobalVoices for \textsc{ces-deu} and
\textsc{eng-ara\_EG} \citep{dettmers2023qlora}. Late recovery helps the matched
domain but degrades both WMT25 OOD metrics (Table~\ref{tab:heal}), so we reject it.

\begin{table}[t]
\centering\small
\setlength{\tabcolsep}{1pt}
\begin{tabular}{@{}lrrrr@{}}
\toprule
& \multicolumn{2}{c}{{\textbf{GlobalVoices}}}
& \multicolumn{2}{c}{{\textbf{WMT25 OOD}}} \\
\cmidrule(lr){2-3}\cmidrule(lr){4-5}
{\scriptsize\textbf{Stage}} & {\scriptsize\textbf{KXXL}$\uparrow$}
& {\scriptsize\textbf{M24XXL}$\downarrow$}
& {\scriptsize\textbf{KXXL}$\uparrow$}
& {\scriptsize\textbf{M24XXL}$\downarrow$} \\
\midrule
i4a8        & 84.21 & 4.282 & 65.05 & 5.837 \\
step=50       & 81.58 & 4.637 & 63.06 & 6.387 \\
step=100      & 81.09 & 4.625 & 62.82 & 6.483 \\
step=400     & 84.61 & 4.123 & 59.94 & 7.234 \\
\bottomrule
\end{tabular}
\caption{QLoRA trajectory for i4a8: 600 held-out GlobalVoices segments versus
the 1121-segment WMT25 OOD set. KXXL=CometKiwi-XXL; M24XXL=MetricX-24-XXL-ref.}
\label{tab:heal}
\end{table}

\begin{table*}[t]
\centering\small
\setlength{\tabcolsep}{5pt}
\renewcommand{\arraystretch}{1.15}
\begin{tabular}{lrrr:rr:rr}
\toprule
& & & & \multicolumn{2}{c}{\textbf{WMT25}} & \multicolumn{2}{c}{\textbf{WMT26 blind}} \\
\cmidrule(lr){5-6}\cmidrule(lr){7-8}
\textbf{System} & \textbf{Disk GB} & \textbf{VRAM GB} & \textbf{tok/s} & \textbf{\kiwixxl{}}$\uparrow$ & \textbf{\metricxref}$\downarrow$ & \textbf{\kiwixxl{}}$\uparrow$ & \textbf{\metricxqe}$\downarrow$ \\
\midrule
bf16    & 23.5 & 23.2 & 4486 & 65.33 & 5.493 & 68.69 & 4.108 \\
i8a16   & 11.8 & 14.4 & 4891 & 65.27 & 5.522 & 68.77 & 4.128 \\
i4a16   & \textbf{6.7} & 9.7 & 5103 & 64.64 & 5.715 & 68.43 & 4.196 \\ \hdashline
f8a8$^{\checkmark,\star}$ & 11.5 & 14.2 & \textbf{6094} & 65.21 & 5.491 & 68.78 & 4.136 \\
i4a8$^{\checkmark}$ & \textbf{6.7} & 9.7 & 5131 & 65.40 & 5.570 & 68.78 & 4.128 \\
f8a8-bok4$^{\checkmark}$ & 12.0 & 12.9 & 1750 & 65.95 & 5.462 & 69.83 & 4.009 \\
i4a8-bok4$^{\checkmark}$ & 6.9 & \textbf{7.5} & 1456 & \textbf{66.32} & 5.467 & \textbf{70.02} & \textbf{3.986} \\
\hdashline
hf-bf16     & 23.5 & -- & -- & 65.28 & 5.529 & 68.65 & 4.156 \\
vllm-bf16   & 23.5 & 26.1 & 4917 & 65.46 & 5.407 & 68.68 & 4.085 \\
vllm-fp8    & 15.3 & 15.8 & 5934 & 65.40 & \textbf{5.378} & 68.68 & 4.096 \\
vllm-nf4    & 8.4 & 11.8 & 3190 & 64.47 & 5.636 & 67.99 & 4.228 \\
\bottomrule
\end{tabular}
\renewcommand{\arraystretch}{1}
\caption{WMT25 development and WMT26 blind-test operating points.
$^{\checkmark}$ marks submissions and $^\star$ the primary. WMT26 uses
reference-free metrics because references were unavailable at the time of
writing. Throughput is hot
steady state and per retained output; best-of-4 disk includes its $0.42$\,GB
selector. hf-bf16 is a quality-only \texttt{transformers} reference.}
\label{tab:trusted-scores}
\end{table*}

\begin{figure*}[t]
\centering
\includegraphics[width=.98\textwidth]{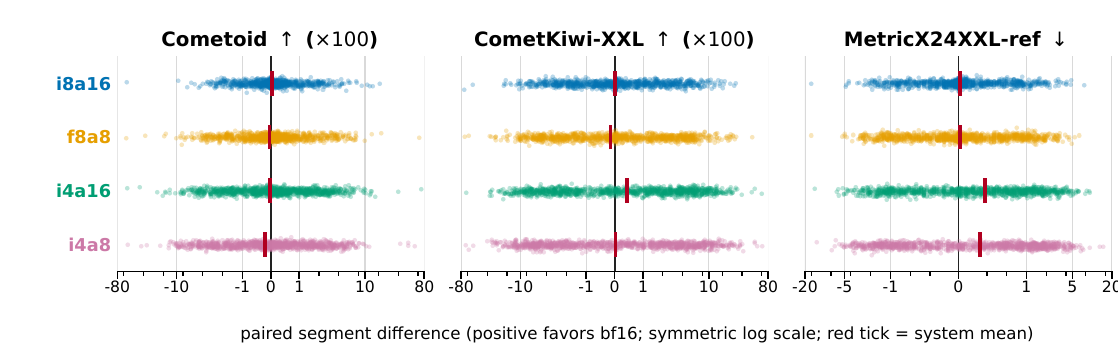}
\caption{Per-segment differences for i8a16, f8a8, i4a16, and i4a8 against bf16
on WMT25 under Cometoid, \kiwixxl{}, and \metricxref{}. Positive values favor
bf16; negative values identify segments where the compressed output scores
better. Red ticks mark means; axes use symmetric log scales.}
\label{fig:seg-diffs}
\end{figure*}


\section{Inference Time Recovery}
\label{sec:inference}

Although i4a8 has a small \metricxref{} residual, compressed systems still beat
bf16 on some segments (Figure~\ref{fig:seg-diffs}). We exploit this variation by
sampling four candidates and selecting the highest-scoring one with the
569M-parameter, reference-free Cometoid QE model \citep{cometoid}. We evaluate
with held-out \kiwixxl{} and \metricxref{}.

\begin{figure}[t]
\centering
\includegraphics[width=.85\linewidth]{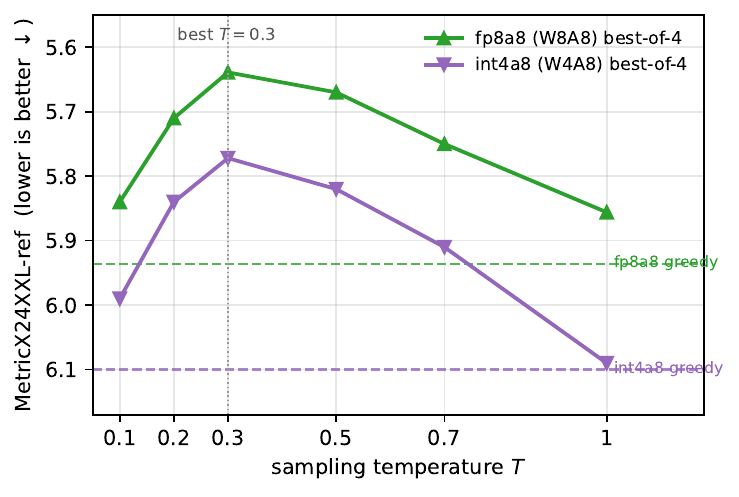}
\caption{Cometoid best-of-4 temperature sweep on WMT25. Solid lines show held-out
\metricxref{} (axis inverted); dashed lines show greedy decoding.}
\label{fig:tempsweep}
\end{figure}

\paragraph{What temperature to use for sampling?}
A single $T{=}1.0$ sample is worse than greedy, and even best-of-4 barely helps.
The WMT25 sweep instead peaks at $T{=}0.3$ for f8a8 and i4a8
(Figure~\ref{fig:tempsweep}), improving \metricxref{} by $0.17$--$0.27$
(Table~\ref{tab:trusted-scores}). We submit these settings as
\texttt{f8a8-bok4} and \texttt{i4a8-bok4}. Their bundled Cometoid selector uses
4-bit weights and int8 embeddings, occupies ${\sim}420$\,MB, and is included in
the reported submission disk size; decoding costs approximately $4\times$ greedy.

\section{Related Work}
\label{sec:related}

\paragraph{Quantization and adaptation.}
Post-training LLM compression includes outlier-aware int8 computation
\citep{dettmers2022llmint8}, weight-only GPTQ and AWQ
\citep{frantar2023gptq,lin2024awq}, and activation quantization such as
SmoothQuant \citep{xiao2023smoothquant}. We instead evaluate disk size, peak VRAM,
H100 throughput, and generated-text quality jointly: at our serving point, int4
weights primarily reduce size, fp8 activations improve speed, and W4A8 combines
both. Calibration-aware quantizers and LoRA/QLoRA depend on representative data
\citep{hu2022lora,dettmers2023qlora}; our GlobalVoices--WMT25 split shows that
in-domain healing can reverse out-of-domain, motivating our OOD gate.

\paragraph{Metric qualification.}
Learned metrics improve over surface-form matching
\citep{rei2020comet,sellam2020bleurt,rei2022cometkiwi,juraska2024metricx24}, but
meta-evaluations show that score magnitudes and system rankings need not match
human preferences \citep{kocmi2021ship,kocmi2024navigating,
freitag-etal-2024-llms,lavie-etal-2025-findings}. Building on paired bootstrap
testing \citep{koehn-2004-statistical}, we use controlled precision changes to
probe metric resolution, quantify ordering stability, and anchor downstream metric
choice in WMT human meta-evaluation. We therefore report resolved differences as
metric residuals, not automatically as human-perceptible losses.

\paragraph{Quality-aware decoding.}
Prior work questions mode-seeking decoding \citep{eikema-aziz-2020-map} and uses
learned metrics for N-best reranking or minimum Bayes-risk decoding
\citep{fernandes-etal-2022-quality,freitag-etal-2022-high}. Reference-free QE makes
selection deployable without references \citep{rei2022cometkiwi,rei2023scaling,
cometoid}. Our cheaper alternative independently scores four low-temperature
samples with a small bundled Cometoid model and is evaluated by held-out arbiters.
Applying it to f8a8 and i4a8 shows that sampling headroom survives quantization.

\section{Conclusion}
\label{sec:conclusion}

Our calibration-free stack removes unused vision/vocabulary components, uses int8
embeddings, and selects f8a8 for speed or i4a8 for size. Int8/fp8 show no resolved
loss; int4 has a small \metricxref{} residual, and QLoRA fails the OOD gate.
Cometoid best-of-4 recovers WMT25 quality and remains competitive on WMT26; we
make no claim that one reranked precision is better than the other.

\section*{Limitations}
Results cover one model, one H100 runtime, and three translation directions, so
speed trade-offs may not transfer to other hardware. QLoRA calibration covers only
two directions. We lack direct human comparisons: WMT24 supplies an external metric
ranking, not judgments on our outputs.

\balance
\bibliography{references}

\end{document}